%% file: main.tex
\documentclass[10pt,twocolumn,letterpaper]{article}

\usepackage[pagenumbers]{cvpr} 

\input{preamble}

\definecolor{cvprblue}{rgb}{0.21,0.49,0.74}
\usepackage[pagebackref,breaklinks,colorlinks,allcolors=cvprblue]{hyperref}

\newcommand{\model}{\textsc{FiReIQ}\xspace}
\newcommand{\data}{\textsc{FiReIQ-Bench}\xspace}

\def\confName{CVPR}
\def\confYear{2027}

\title{FiReIQ: Field-Level Confidence Estimation with Reasoning and Image Quality in Document Extraction}

\author{
Xiao Liu \quad
Yage Wang \quad
Qi Yu \quad
Giri Iyengar \\[0.5em]
Capital One
}

\begin{document}
\maketitle
\input{sec/0_abstract}

\input{sec/1_intro}
\input{sec/2_related}
\input{sec/3_method}

\input{sec/4_data}

\input{sec/5_experiments}

\input{sec/6_analysis}
\input{sec/7_conclusion}
{
    \small
    \bibliographystyle{ieeenat_fullname}
    \bibliography{main}
}

\input{sec/X_suppl}

\end{document}

%% file: preamble.tex
\usepackage{booktabs}  
\usepackage{multirow}
\usepackage[table]{xcolor}  

\usepackage{graphicx}  
\usepackage{listings}

%% file: sec/0_abstract.tex
\begin{abstract}
Vision-Language Models (VLMs) have demonstrated strong
performance in structured document extraction, yet
their deployment in high-stakes applications requires
reliable confidence estimates for individual fields.
Existing uncertainty estimators often aggregate signals
across an entire generation, obscuring field-specific
reasoning uncertainty while overlooking visual
degradations that may lead to confident extraction errors.
To address these limitations, we introduce \model,
a field-level confidence estimation framework that
combines field-aligned reasoning signals from a single
VLM generation with local and global image-quality
features through quality-conditioned visual fusion.
To facilitate systematic evaluation, we further introduce
\data, a benchmark spanning financial receipts and
legal contracts. Experiments with two reasoning VLMs
show that \model consistently improves confidence
discrimination, selective prediction, and probabilistic
accuracy over existing estimators. These gains extend
to private financial service data, where \model improves
AUROC by 6.8\% and coverage at
95\% precision by 10.6\% over an
existing baseline. Together, these results highlight
the potential of \model to reduce manual review
while enabling more reliable document automation.
\end{abstract}

%% file: sec/1_intro.tex
\section{Introduction}
\label{sec:intro}

Recent advances in Vision-Language Models (VLMs) have substantially
improved structured information extraction from visual documents~\cite{donut, qwen25_vl},
enabling applications such as financial document processing,
automated data entry, and legal contract analysis. Despite these
advances, extraction errors remain a critical concern in high-stakes
applications, where an incorrect numerical value or misidentified
legal entity can propagate into downstream decisions. Therefore, reliable deployment  requires not only accurate extraction but also
field-level confidence estimates that identify potentially incorrect
predictions and enable selective human review~\cite{doc_ie_failure_prediction, selective_risk_control, selective_classification}.

However, reliable confidence estimation for multi-field document
extraction presents two key challenges. First, uncertainty must be
attributed to individual fields rather than the entire generation.
A single extraction typically produces multiple fields with varying
levels of correctness and uncertainty. Existing uncertainty estimation
methods often rely on consistency across repeated generations
~\cite{semantic_entropy, kle}, introducing additional inference costs.
Although probability-based and model-probing approaches offer more
efficient alternatives~\cite{ptrue, llm_internal_state} while they are chanllenging on extending to field-level and aggregating uncertainty over
the entire generation can obscure the uncertainty associated with
individual fields~\cite{claim_dilution}. Recent
methods such as ExtractConf~\cite{extractconf}, address field-level
confidence by combining multiple extraction calls with external OCR
signals. However, obtaining field-specific uncertainty directly
from a single generation, without additional extraction calls or
external OCR, remains an open challenge.

Second, uncertainty inferred from the model's internal signal may not fully reflect the reliability of the underlying visual evidence. Document extraction is sensitive to image degradation, such as blur, noise, and low contrast~\cite{diqa_survey, wilddoc}, which can lead VLMs to produce confident but incorrect predictions~\cite{seeing_without_looking, visual_quality_paradox}. Existing VLM confidence methods primarily rely on semantic visual representations or grounding signals~\cite{reliable_vqa, halp, fuse, bicr}, which do not explicitly characterize image degradation. Although recent document-specific methods incorporate image-quality cues, such as the local sharpness measurements used by ExtractConf~\cite{extractconf}, they typically treat these measurements as auxiliary classifier features. However, image degradation varies in both type and location, making a single quality metric insufficient to characterize visual reliability. Moreover, simply treating image quality as an additional feature may limit its ability to influence how visual information is represented. These limitations motivate richer image-quality representations and more effective mechanisms for integrating quality information into confidence estimation.

To address these challenges, we introduce \model, a field-level confidence estimation framework that integrates reasoning uncertainty with explicit document image-quality information. First, we leverage structured decoding~\cite{vllm_structured_outputs} to obtain field-specific reasoning excerpts and their associated token probabilities from a single VLM generation. We then construct complementary reasoning features that capture both token-level uncertainty patterns and semantic cues expressed in the reasoning text. Second, to account for unreliable visual evidence, we develop an image-quality representation that characterizes multiple degradation factors at both local and global scales. Rather than treating these quality measurements as auxiliary features, we employ Feature-wise Linear Modulation (FiLM)~\cite{film} to adapt a pretrained visual representation according to the measured document quality. The resulting image-quality-aware representation is fused with field-specific reasoning features to predict the correctness probability of each extracted field. Finally, we calibrate these probabilities to support selective field acceptance and human review.

To support systematic evaluation, we introduce
\data, a benchmark for field-level confidence
estimation in multi-field document extraction.
Built upon CORD~\cite{cord} and CUAD~\cite{cuad},
\data covers financial receipts and legal contracts,
pairing extracted fields with field-specific reasoning
traces, document visual features, and correctness labels.
We also preserve raw reasoning token probability distributions
to provide a unified representation of reasoning states
for future research on uncertainty estimation.
We evaluate \model using two reasoning VLMs,
Gemma 26B~\cite{gemma_backbone} and Nemotron 3.5 Super VL 120B~\cite{nemotron_backbone}.
Across both backbones, \model consistently improves
confidence discrimination and selective prediction
over existing estimators. In particular, incorporating
quality-conditioned visual features reduces overall
AURC~\cite{bias_reduced_uncertainty} by 9.4\% on Gemma and 21.2\% on Nemotron
compared with the corresponding model without
explicit image-quality conditioning.
Beyond public benchmarks, we further evaluate
\model on private financial service data
covering high-stakes financial workflows.
Compared with an existing baseline,
\model improves AUROC by 6.8 percentage points
and automatic coverage at 95\% precision by
10.6 percentage points. These findings demonstrate
the complementary value of field-specific reasoning
and image-quality awareness for reliable document
extraction, with the potential to reduce manual
review costs in high-stakes financial workflows.

\noindent Our main contributions are summarized as follows:
\begin{itemize}
    \item We introduce \model, a single-generation
    confidence estimation framework that attributes
    reasoning uncertainty to individual extracted fields.
    By leveraging field-specific reasoning text and
    token probabilities, it estimates extraction
    reliability without repeated sampling or external OCR.

    \item We propose a quality-aware confidence model
    that combines field-level reasoning uncertainty
    with local and global image-quality measurements.
    Through FiLM-based conditioning, image quality
    adaptively modulates visual representations to
    improve confidence discrimination and selective
    prediction.

   \item We introduce \data, a benchmark spanning
    financial receipts and legal contracts that preserves
    field-specific reasoning traces and raw token probability
    distributions to support future research on uncertainty
    estimation. Experiments across two reasoning VLMs and private financial
data demonstrate improved confidence estimation and
reliable document automation.
\end{itemize}

%% file: sec/2_related.tex
\begin{figure*}[t]
    \centering
    \includegraphics[width=\textwidth]{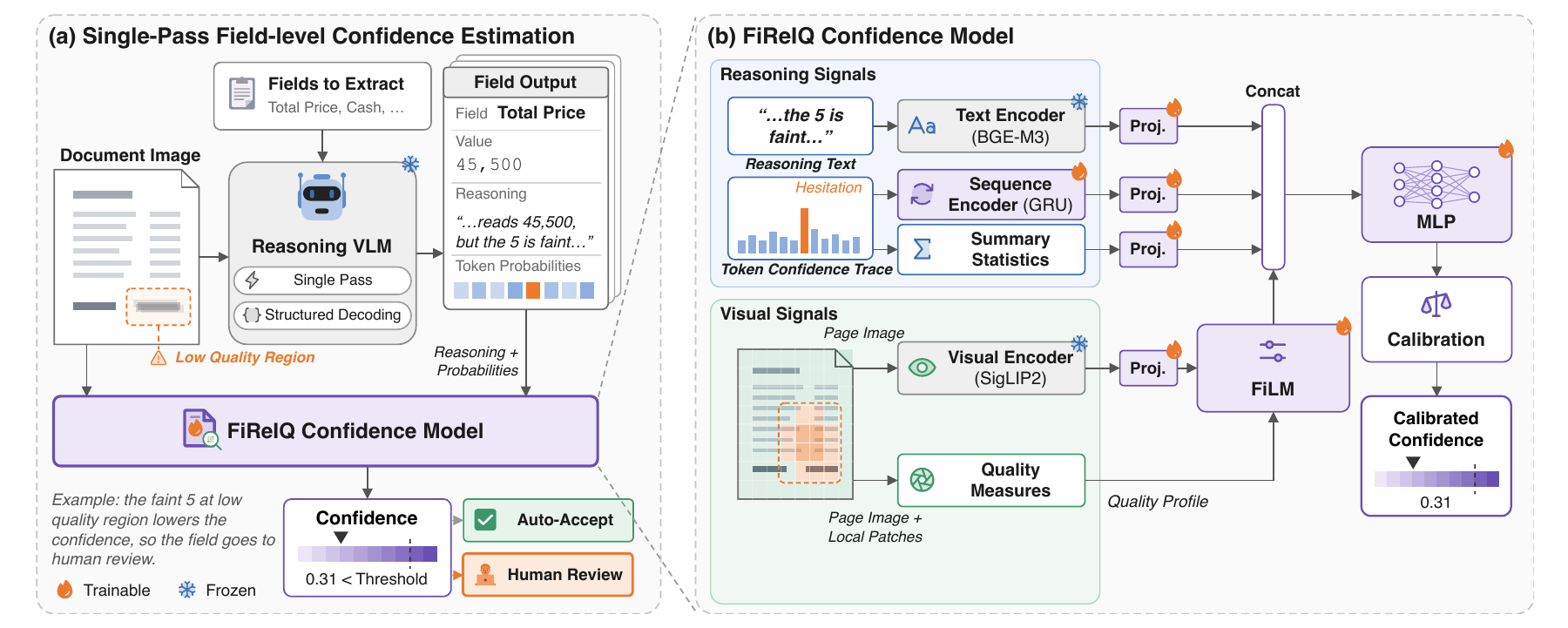}
    \caption{Overview of \model. (a) A reasoning VLM extracts all
    fields in a single pass with structured decoding and returns, for
    each field, its value, an excerpt of its reasoning, and the token
    probabilities. The confidence model combines these with the page
    image to produce a calibrated confidence, which decides whether
    the field is accepted automatically or sent to human review.
    (b) Inside the confidence model, per-field reasoning signals and
    per-page image signals are encoded and projected. The image-quality
    profile modulates the visual features through FiLM before fusion,
    and an MLP followed by calibration yields the final confidence.}
    \label{fig:pipeline}
\end{figure*}

\section{Related Work}
\label{sec:related}

\subsection{Uncertainty Estimation in LLMs and VLMs}
\label{sec:related:uq}

Uncertainty estimation for large language models (LLMs)
typically relies on semantic consistency across sampled
generations~\cite{semantic_entropy, kle}, token probabilities
or self-evaluation~\cite{ptrue}, or verbalized
confidence~\cite{verbalized_confidence}. For reasoning
models, recent work extracts confidence from the reasoning
trace through its length~\cite{trace_length}, token-level
confidence trajectories~\cite{confidence_geometry,
entropy_trajectory}, intermediate
answers~\cite{beyond_last_answer}, or step-level
confidence~\cite{recurrent_confidence_chain,
stepwise_attribution}, assigning one confidence to each
response or step. Claim-level
calibration~\cite{claim_dilution} shows that such confidence
is too coarse for generations with multiple claims, but
scores each claim through repeated sampling and
self-verification.

Vision-language models (VLMs) are also often miscalibrated
and overconfident~\cite{overconfidence_key,
unveiling_uncertainty, seeing_is_believing}. Beyond
abstention in visual question answering~\cite{reliable_vqa},
recent VLM methods contrast predictions with and without
the image~\cite{vauq, bicr}, weight token uncertainty by
visual grounding~\cite{vig_tuq}, calibrate confidence with
a grounding model~\cite{grounding_calibration}, fuse
aleatoric and epistemic uncertainty~\cite{fuse}, or probe
internal representations for hallucination
risk~\cite{halp}. However, these semantic
and grounding signals do not explicitly characterize image
degradation, and these methods do not attribute confidence
to individual fields within a generation.

\subsection{Confidence in Document Extraction}
\label{sec:related:doc}

In document extraction, ConfBERT~\cite{confbert} uses OCR
confidence to detect post-OCR errors. For LLM-based extraction,
CONSTRUCT~\cite{construct} scores the trustworthiness of
each field in a structured output, per-field selective risk
control~\cite{selective_risk_control} studies statistical
guarantees for accepted fields, and
ConfBench~\cite{confbench} benchmarks VLM confidence under
controlled document degradations. Most related to our work,
ExtractConf~\cite{extractconf} fuses disagreement between
two extraction calls, token probabilities, OCR confidence,
layout, and local Laplacian sharpness in a gradient-boosted
classifier. In contrast, \model derives field-level
confidence from the reasoning of a single generation
without external OCR.

Image quality also affects VLM reliability. Common
corruptions reduce VLM accuracy~\cite{corruption_robustness,
vlm_robustbench} and impair verbalized
uncertainty~\cite{corrupted_verbalized, stated_vs_internal},
while higher perceptual quality does not necessarily improve
performance~\cite{visual_quality_paradox}. VLMs can also be
insensitive to the loss of fine-grained visual
evidence~\cite{seeing_without_looking}. Document image
quality assessment estimates perceptual quality or OCR
accuracy~\cite{diqa_survey, kang2014diqa, smartdoc_qa,
dociq}, and image quality has been used to route documents
before extraction~\cite{preinference_routing}. In contrast,
\model measures multiple local and global degradation
factors and uses them to condition visual representations
through FiLM~\cite{film} for field-level confidence
estimation.

%% file: sec/3_method.tex
\section{\model}
\label{sec:method}

In this section, we present \model, our field-level confidence
estimation framework. \model first extracts field-aligned reasoning
traces and token probabilities from a single generation and constructs
reasoning-based uncertainty features for each field. To capture errors
that may not be reflected in the model's reasoning, we additionally
incorporate visual features derived from the document image.
These complementary signals are integrated through a
quality-conditioned FiLM network to predict field-level correctness
probabilities, which are subsequently calibrated for selective
prediction. Figure~\ref{fig:pipeline} illustrates the overall pipeline.

\subsection{Single-Pass Field Reasoning Extraction}
\label{sec:method:extraction}

Prior work suggests that a model's reasoning trajectory contains richer
uncertainty signals than its final answer alone~\cite{claim_dilution,
semantic_entropy,beyond_last_answer}. Motivated by this observation, \model captures
field-specific reasoning and token probabilities within a single
generation, without repeated sampling for confidence estimation
or an additional model invocation.

Specifically, we use vLLM's structured decoding~\cite{vllm,vllm_structured_outputs}
to constrain the model's response to a predefined JSON schema. For each target
field, the schema includes an extracted value and a \texttt{reasoning} field.
We instruct the model to populate the latter with verbatim excerpts from its
raw reasoning trace, rather than paraphrased explanations. We also record
token-level probability information during generation, allowing each reasoning
excerpt to be associated with its corresponding token probabilities. This
produces machine-readable field extractions, field-aligned reasoning excerpts,
and their token probabilities within a single generation. The complete output
schema is provided in Supplementary Material~\ref{sec:supp:schema}.

\subsection{Field-Level Reasoning Features}
\label{sec:method:trace}

Given the field-specific reasoning text and token probabilities obtained
above, we construct two complementary types of uncertainty features:
reasoning probability features that characterize token-level uncertainty,
and semantic features that capture uncertainty expressed in the reasoning
text.

\paragraph{Reasoning Probability Features.}
For each field $f$, let $T_f$ denote the number of tokens in its
reasoning trace. We first compute a token-level uncertainty score
using the mean negative log probability of the top-$k$ candidate
tokens~\cite{deepconf}:
\begin{equation}
    c_i^{f} = -\frac{1}{k}\sum_{j=1}^{k}\log p_{i,j}^{f},
    \label{eq:token_conf}
\end{equation}
where $p_{i,j}^{f}$ denotes the probability of the $j$-th most
likely token at position $i$ in the reasoning trace of field $f$.
The resulting uncertainty sequence is
$\mathbf{c}^{f} = [c_1^{f}, \ldots, c_{T_f}^{f}]$.

We derive two complementary representations from this sequence.
First, we retain the last $L$ uncertainty scores and encode them
using a gated recurrent unit (GRU)~\cite{gru}:
\begin{equation}
    \mathbf{h}_{\mathrm{seq}}^{f}
    = \operatorname{GRU}\left(
    \mathbf{c}_{\mathrm{tail}}^{f}
    \right),
    \label{eq:seq_feat}
\end{equation}
where $\mathbf{c}_{\mathrm{tail}}^{f}$ denotes the last $L$
scores, with padding applied when $T_f<L$. This representation
captures sequential variations in uncertainty.

Second, we compute a statistical feature vector over the same
retained uncertainty scores:
\begin{equation}
    \mathbf{h}_{\mathrm{stat}}^{f}
    = \operatorname{Stats}\left(\mathbf{c}_{\mathrm{tail}}^{f}\right)
    \in \mathbb{R}^{12},
    \label{eq:stat_feat}
\end{equation}
where $\operatorname{Stats}(\cdot)$ computes twelve statistics
over the non-padded scores: sequence length, mean, standard
deviation, minimum, maximum, median, first and last values,
mean of the final five values, linear trend, and the 25th and
75th percentiles. These statistics summarize the distribution
and variation of uncertainty across the final $L$ reasoning
tokens, complementing the GRU-based representation.

\paragraph{Reasoning Semantic Features.}
Token probabilities may not fully capture uncertainty expressed
through language, such as hedging, ambiguity, or competing
interpretations. We therefore encode the field-specific reasoning
text $r^{f}$ using a pretrained BGE-M3 encoder~\cite{bge_m3}:
\begin{equation}
    \mathbf{h}_{\mathrm{text}}^{f}
    = \operatorname{BGE\text{-}M3}\left(r^{f}\right).
    \label{eq:text_feat}
\end{equation}
This semantic representation complements the probability-based
features by capturing linguistic cues of uncertainty.

Together, $\mathbf{h}_{\mathrm{seq}}^{f}$,
$\mathbf{h}_{\mathrm{stat}}^{f}$, and
$\mathbf{h}_{\mathrm{text}}^{f}$ form the field-level reasoning
features used by \model.

\subsection{Image-Quality-Aware Visual Features}
\label{sec:method:quality}

While reasoning-derived features capture uncertainty expressed during
generation, they may fail to identify errors caused by degraded visual
evidence. To address this limitation, \model explicitly
incorporates document image quality alongside a pretrained visual
representation.

\paragraph{Visual Semantic Features.}
Given a document image $I$, we extract a page-level visual embedding
using a frozen base SigLIP2 encoder~\cite{siglip2}:
\begin{equation}
    \mathbf{h}_{\mathrm{vis}} = \operatorname{SigLIP2}(I).
    \label{eq:visual_feat}
\end{equation}
This embedding provides general visual context, while the image-quality
features introduced below explicitly characterize visual degradation.

\paragraph{Image Quality Features.}
To capture both local and global image degradation, we construct a
spatially resolved no-reference quality representation. Specifically,
we divide each document image into an $8\times8$ grid of tiles
$\{I_t\}_{t=1}^{64}$ and compute six quality metrics for each tile:
edge acutance, frequency-domain sharpness, image entropy, perceptual
blur, contrast, and estimated noise level. Let $\phi(\cdot)$ denote
the resulting six-dimensional quality descriptor. We concatenate
the tile-level descriptors with the same metrics computed globally:
\begin{equation}
    \mathbf{q}
    = [\phi(I_1); \ldots; \phi(I_{64}); \phi(I)]
    \in \mathbb{R}^{390}.
    \label{eq:quality_feat}
\end{equation}
Unlike global quality measurements alone, this representation
preserves spatial variations in document quality, capturing localized
degradations that may affect extraction reliability. Definitions of the
six quality metrics are provided in Supplementary
Material~\ref{sec:supp:quality_metrics}.

Both representations are computed at the page level and shared across
fields. In the following section, we use the proposed quality vector
$\mathbf{q}$ to condition the visual embedding $\mathbf{h}_{\mathrm{vis}}$,
producing an image-quality-aware representation that is subsequently
fused with field-specific reasoning features.

\subsection{Quality-Conditioned Confidence Model}
\label{sec:method:confmodel}

Given the field-specific reasoning features and document visual
representations, \model integrates these complementary signals
to predict the correctness of each extracted field. Specifically,
we employ Feature-wise Linear Modulation (FiLM)~\cite{film} to incorporate
image-quality information into the visual representation before
fusing it with the reasoning features.

\paragraph{Image Quality Conditioning.}
We first project the visual embedding $\mathbf{h}_{\mathrm{vis}}$ into a
latent representation using a learnable projection block
$P_{\mathrm{vis}}(\cdot)$:
\begin{equation}
    \mathbf{z}_{\mathrm{vis}}
    = P_{\mathrm{vis}}(\mathbf{h}_{\mathrm{vis}}).
\end{equation}
Rather than directly concatenating the quality vector $\mathbf{q}$
with other features, we use it to generate feature-wise scaling and
shifting parameters through a lightweight conditioning network.
This design allows \model to adaptively adjust the contribution of
visual features according to the document's image quality, potentially
reducing reliance on unreliable visual evidence while preserving
informative features. The resulting image-quality-aware representation is
\begin{equation}
    \widetilde{\mathbf{z}}_{\mathrm{vis}}
    = \left(1+\gamma(\mathbf{q})\right)
    \odot \mathbf{z}_{\mathrm{vis}} + \beta(\mathbf{q}),
    \label{eq:image_film}
\end{equation}
where $\gamma(\cdot)$ and $\beta(\cdot)$ are learnable functions
that produce the FiLM scaling and shifting parameters, respectively,
and $\odot$ denotes element-wise multiplication.

\paragraph{Multimodal Fusion.}
We independently project the three field-specific reasoning features
using learnable projection blocks $P_{\mathrm{seq}}$,
$P_{\mathrm{stat}}$, and $P_{\mathrm{text}}$. The projected features
are concatenated with the image-quality-aware visual representation:
\begin{equation}
\begin{aligned}
    \mathbf{z}^{f} = \big[
    &P_{\mathrm{seq}}(\mathbf{h}_{\mathrm{seq}}^{f});
    P_{\mathrm{stat}}(\mathbf{h}_{\mathrm{stat}}^{f});\\
    &P_{\mathrm{text}}(\mathbf{h}_{\mathrm{text}}^{f});
    \widetilde{\mathbf{z}}_{\mathrm{vis}}
    \big].
\end{aligned}
\label{eq:multimodal_fusion}
\end{equation}
Each projection block consists of a linear layer, layer normalization~\cite{layer_norm},
ReLU activation~\cite{relu}, and dropout~\cite{dropout}. The fused representation is then passed
through a multilayer perceptron (MLP) and a linear prediction head to
produce a field-level correctness logit:
\begin{equation}
    s^{f} = \mathbf{w}^{\top}
    \operatorname{MLP}(\mathbf{z}^{f}) + b.
    \label{eq:confidence_logit}
\end{equation}
The confidence model is trained using binary cross-entropy loss with
positive-class weighting to address label imbalance.
Detailed network configurations and training hyperparameters
are provided in Supplementary Material~\ref{sec:method:impl}.

\paragraph{Probability Calibration.}
To obtain calibrated correctness probabilities, we apply a post hoc
affine temperature transformation~\cite{platt_scaling,temperature_scaling} to the predicted logits:
\begin{equation}
    \hat{p}^{f}
    = \sigma\left(\frac{s^{f}}{T}+b_{\mathrm{cal}}\right),
    \label{eq:confidence_calibration}
\end{equation}
where $\sigma(\cdot)$ is the sigmoid function, and the shared temperature
$T>0$ and bias $b_{\mathrm{cal}}$ are optimized on a held-out calibration
set by minimizing negative log-likelihood. This monotonic transformation
preserves the ranking of field predictions while adjusting their
probability values. The calibrated probabilities are then used with a
shared threshold to determine whether each field is automatically
accepted or referred for human review.

%% file: sec/4_data.tex
\section{\data}
\label{sec:data}

To train and evaluate \model, we construct \data, a benchmark for
field-level confidence estimation spanning two representative
high-stakes document extraction tasks: numerical value extraction
from financial documents and entity extraction from legal contracts.
These tasks present complementary challenges, from precise numerical
recognition to identifying contract-specific entities within complex
document structures.

\subsection{Source Documents}
\label{sec:data:sources}

We select two public datasets representing complementary document
extraction scenarios.

\noindent\textbf{CORD v2}~\cite{cord} is a receipt understanding dataset
containing real-world receipt images with structured annotations.
We focus on 24 transaction-related fields, such as item names, quantities,
prices, and totals. The diverse layouts and image conditions pose
challenges for accurate numerical extraction, where even a single
incorrect digit can affect correctness.

\noindent\textbf{CUAD}~\cite{cuad} is a legal contract understanding
dataset comprising commercial contracts with expert annotations
across 41 categories. We select 8 fields, including contracting
parties, governing law, and agreement and expiration dates.
These contracts require identifying relevant entities and attributes
within lengthy and structurally complex documents. We render the
contracts into page images and align annotations with their
corresponding pages.

For each dataset, we define a domain-specific extraction schema.
The complete list of extraction fields and their definitions is
provided in Supplementary Material~\ref{sec:supp:fields}.

\input{tables/data_stats}

\subsection{Data Collection and Labeling}
\label{sec:data:labeling}

We construct \data using the extraction pipeline described in
Section~\ref{sec:method:extraction} with two reasoning VLMs:
Gemma 26B~\cite{gemma_backbone} and Nemotron 3.5 Super VL 120B~\cite{nemotron_backbone}. For each model,
we generate three independent completions per page. Each field
in a completion constitutes one example, consisting of its
extracted value, field-specific reasoning features, and
page-level visual features. Confidence features are computed
independently for each completion, without aggregating
information across generations.

We assign each example a binary correctness label by comparing
its extracted value against the corresponding ground truth.
We employ field-type-specific matching criteria to accommodate
formatting variations while preserving extraction correctness.
Detailed matching criteria are provided in Supplementary
Material~\ref{sec:supp:labeling}.

Table~\ref{tab:data_stats} summarizes the statistics of \data
across the two VLMs. Extraction accuracy varies substantially
between domains, with both models achieving over 90\% accuracy
on CORD but below 60\% on CUAD, reflecting different levels
of extraction difficulty. We partition the data at the document
level to prevent leakage across splits, reserving separate
training, calibration, and test sets. The calibration split
is used exclusively for probability calibration and confidence
threshold selection. Further details of the partitioning and
calibration protocol are provided in Supplementary
Material~\ref{sec:supp:main_protocol}.

%% file: tables/data_stats.tex
\begin{table}[!t]
\centering
\caption{Statistics of the field-level confidence datasets.
Train, Cal., and Test denote the number of field-level
examples in each split. Acc. (\%) denotes extraction accuracy.}
\label{tab:data_stats}
\small
\scalebox{0.8}{%
\begin{tabular}{llrrrrr}
\toprule
\textbf{Dataset} & \textbf{VLM} & \textbf{\# Pages}
& \textbf{Train} & \textbf{Cal.} & \textbf{Test}
& \textbf{Acc.} \\
\midrule
CORD & Gemma 26B     & 1,000 & 22,132 & 2,787 & 2,722 & 92.3 \\
     & Nemotron 120B & 776   & 13,271 & 1,709 & 1,671 & 89.9 \\
\midrule
CUAD & Gemma 26B     & 198   & 3,926  & 639   & 494   & 59.2 \\
     & Nemotron 120B & 198   & 5,182  & 755   & 450   & 58.4 \\
\bottomrule
\end{tabular}
}
\end{table}

%% file: sec/5_experiments.tex
\section{Experiments}
\label{sec:experiments}

We evaluate \model on \data to assess its effectiveness in
estimating field-level correctness and supporting selective
document extraction. We compare \model with conventional
confidence estimators and learned reasoning-based and
multimodal baselines across two document domains and
two reasoning VLMs.

\subsection{Experimental Setup}
\label{sec:experiments:setup}

We evaluate two reasoning
VLMs, Gemma 26B~\cite{gemma_backbone} and Nemotron 3.5 Super VL 120B~\cite{nemotron_backbone},
using the field-level examples constructed in
Section~\ref{sec:data}. Each confidence estimator is
evaluated on the same field predictions and correctness
labels. For data partitioning, we use
document-level training, calibration, and test splits instead of page-level
to prevent information leakage across partitions.
We include more implementation details, including feature preprocessing
and optimization settings in Supplementary
Material~\ref{sec:method:impl}.

\subsection{Baselines}
\label{sec:experiments:baselines}

We compare \model with three groups of confidence estimators:

\begin{itemize}
    \item \textbf{Training-free Estimators.}
    Sample Agreement measures field-level agreement among
    three independent completions~\cite{self_consistency, extractconf}. Answer LogProb and
    Reasoning LogProb aggregate token probabilities over
    the extracted field value~\cite{confbench, extractconf} and field-specific reasoning,
    respectively.

    \item \textbf{Reasoning-based Estimators.}
    Reasoning Trajectory learns from sequential and
    statistical uncertainty features, while Reasoning + Text
    additionally incorporates semantic reasoning embeddings
    from BGE-M3~\cite{bge_m3}.

    \item \textbf{Multimodal Estimators.}
    Reasoning + Visual combines reasoning features with
    pretrained SigLIP2~\cite{siglip2} image embeddings without explicitly
    modeling image quality. Our full \model further conditions
    the visual representation on local and global image-quality
    measurements through FiLM~\cite{film}.
\end{itemize}

\noindent All learned baselines are trained independently using
the same training protocol and data splits, with only
their enabled feature branches differing. Additional
answer-probability baselines and implementation details
are provided in Supplementary Material~\ref{sec:supp:main_comparison}.

\subsection{Evaluation Metrics}
\label{sec:experiments:metrics}

We evaluate confidence estimation from four perspectives:

\begin{itemize}
    \item \textbf{Selective Prediction.}
    Our primary metric is the area under the
    risk--coverage curve (AURC)~\cite{selective_classification, bias_reduced_uncertainty}, where risk denotes
    the error rate among accepted field predictions
    at a given coverage level. Lower AURC indicates
    better selective prediction.

    \item \textbf{Discrimination.}
    We report the area under the receiver operating
    characteristic curve (AUROC)~\cite{roc_analysis, misclassification_baseline} to measure how well
    confidence scores distinguish correct from
    incorrect field extractions.

    \item \textbf{Automation.}
    We report coverage at 95\% and 98\% precision
    (Cov@95/98), measuring the proportion of
    fields that can be automatically accepted while
    satisfying the corresponding precision requirement~\cite{selective_classification}.
    These thresholds represent the minimum acceptable
    and preferred precision levels for practical
    applications, respectively. Higher coverage
    indicates greater automation.

    \item \textbf{Calibration and Probability Quality.}
    We report expected calibration error (ECE)~\cite{ece_naeini, temperature_scaling} to
    measure the agreement between predicted confidence
    and empirical accuracy, and the Brier score~\cite{brier_score} to
    assess probabilistic accuracy. Lower values
    are better for both metrics.
\end{itemize}

\noindent Formal definitions of all evaluation metrics are
provided in Supplementary Material~\ref{sec:supp:main_protocol}.





\input{tables/main_exp}

\subsection{Main Results}
\label{sec:experiments:main}

Table~\ref{tab:main_results} summarizes the comparison
across CORD and CUAD using both reasoning VLMs.
Complete results for all baselines, including additional
answer-probability estimators, are reported in
Tables~\ref{tab:supp:cord_full} and~\ref{tab:supp:cuad_full}
of Supplementary Material~\ref{sec:supp:baseline_results}.
\\

\noindent\textbf{\model consistently improves confidence
ranking across document domains and VLM backbones.}
\model achieves the lowest AURC and highest AUROC
across all four settings. Compared with
Reasoning + Visual, AURC decreases from 0.1514
to 0.1222 on Gemma CUAD and from 0.0273 to
0.0187 on Nemotron CORD, corresponding to relative
reductions of 19.3\% and 31.5\%, respectively.
These consistent improvements indicate that explicit
image-quality modeling provides complementary
information beyond reasoning uncertainty and
pretrained visual representations.
\\

\noindent\textbf{\model enables greater automation
under stringent precision requirements.}
At 98\% precision, \model increases coverage
from 0.6468 to 0.8684 on Gemma CORD and from
0.2671 to 0.6900 on Nemotron CORD compared with
Reasoning + Visual. Improvements are also evident
on the more challenging CUAD dataset, where
Cov@98 increases from 0.0227 to 0.1935 with
Gemma and from 0.0143 to 0.0797 with Nemotron.
These gains demonstrate the practical value
of confidence estimation for identifying
reliable fields suitable for automatic acceptance.
\\


\noindent\textbf{\model improves probabilistic accuracy
while maintaining competitive calibration.}
Beyond discrimination and selective coverage, \model
achieves the lowest Brier score across all four
dataset--VLM settings. As shown in
Table~\ref{tab:main_results}, it reduces Brier
score on CUAD by 15.9\% with Gemma and 16.2\%
with Nemotron relative to Reasoning + Visual.
It also achieves the lowest ECE on both Nemotron
settings, including a 52.1\% reduction on CUAD,
while remaining competitive on Gemma. These results
demonstrate improved probabilistic accuracy, although
calibration gains vary across backbones.

%% file: tables/main_exp.tex
\begin{table*}[t]
\centering
\caption{Comparison of field-level confidence estimation methods
on \data. AURC measures selective prediction risk, AUROC
measures confidence discrimination, Cov@95 and Cov@98 denote
coverage at 95\% and 98\% precision, respectively, while
ECE and Brier evaluate probability quality.
Best results within each dataset and backbone are in bold.}
\label{tab:main_results}

\footnotesize
\setlength{\tabcolsep}{3pt}
\renewcommand{\arraystretch}{1.12}

\begin{tabular}{l|cccccc|cccccc}
\toprule
& \multicolumn{6}{c|}{\textbf{Gemma 26B}}
& \multicolumn{6}{c}{\textbf{Nemotron 3.5 Super VL 120B}} \\
\cmidrule(lr){2-7}
\cmidrule(lr){8-13}
\textbf{Method}
& AURC$\downarrow$
& AUROC$\uparrow$
& Cov@95$\uparrow$
& Cov@98$\uparrow$
& ECE$\downarrow$
& Brier$\downarrow$
& AURC$\downarrow$
& AUROC$\uparrow$
& Cov@95$\uparrow$
& Cov@98$\uparrow$
& ECE$\downarrow$
& Brier$\downarrow$ \\
\midrule

\rowcolor{gray!15}
\multicolumn{13}{c}{\textbf{CORD}} \\
\midrule

Sample Agreement
& 0.0367 & 0.6648 & 0.8635 & 0.2115 & 0.0264 & 0.0641
& 0.0503 & 0.7375 & 0.7115 & 0.0156 & 0.0254 & 0.0873 \\

Answer LogProb
& 0.0544 & 0.4960 & \textbf{0.9869} & 0.0018 & \textbf{0.0179} & 0.0489
& 0.1083 & 0.4324 & 0.0000 & 0.0000 & 0.0384 & 0.0879 \\

Reasoning LogProb
& 0.0545 & 0.5593 & 0.2661 & 0.0419 & 0.0271 & 0.0651
& 0.1051 & 0.4923 & 0.0030 & 0.0030 & 0.0374 & 0.1008 \\

\midrule

Reasoning Trajectory
& 0.0334 & 0.7316 & 0.9164 & 0.1832 & 0.0210 & 0.0583
& 0.0614 & 0.6752 & 0.4172 & 0.0355 & 0.0310 & 0.0919 \\

Reasoning + Text
& 0.0168 & 0.8497 & 0.9774 & 0.7086 & 0.0223 & 0.0434
& 0.0342 & 0.8487 & 0.9178 & 0.0041 & 0.0307 & 0.0505 \\

\midrule

Reasoning + Visual
& 0.0173 & 0.8414 & 0.9774 & 0.6468 & 0.0260 & 0.0456
& 0.0273 & 0.8592 & 0.9122 & 0.2671 & 0.0287 & 0.0558 \\

\midrule

\textbf{\model (Ours)}
& \textbf{0.0166}
& \textbf{0.8736}
& 0.9839
& \textbf{0.8684}
& 0.0188
& \textbf{0.0387}
& \textbf{0.0187}
& \textbf{0.8907}
& \textbf{0.9539}
& \textbf{0.6900}
& \textbf{0.0197}
& \textbf{0.0460} \\

\midrule

\rowcolor{gray!15}
\multicolumn{13}{c}{\textbf{CUAD}} \\
\midrule

Sample Agreement
& 0.3998 & 0.4640 & 0.0076 & 0.0076 & \textbf{0.0902} & 0.2499
& 0.3842 & 0.5067 & 0.0038 & 0.0038 & 0.0898 & 0.2467 \\

Answer LogProb
& 0.2464 & 0.4292 & 0.0158 & 0.0090 & 0.1175 & 0.1824
& 0.3506 & 0.3877 & 0.0042 & 0.0042 & 0.0564 & 0.2024 \\

Reasoning LogProb
& 0.3057 & 0.6431 & 0.0000 & 0.0000 & 0.1494 & 0.2323
& 0.3743 & 0.4857 & 0.0000 & 0.0000 & 0.1164 & 0.2552 \\

\midrule

Reasoning Trajectory
& 0.2162 & 0.7316 & 0.0095 & 0.0095 & 0.1689 & 0.2201
& 0.2782 & 0.6150 & 0.0129 & 0.0128 & 0.1142 & 0.2390 \\

Reasoning + Text
& 0.1557 & 0.8662 & 0.2010 & 0.0327 & 0.1225 & 0.1434
& 0.1468 & 0.8649 & 0.0903 & 0.0186 & 0.0739 & 0.1281 \\

\midrule

Reasoning + Visual
& 0.1514 & 0.8684 & 0.2397 & 0.0227 & 0.1341 & 0.1621
& 0.1620 & 0.8429 & 0.0456 & 0.0143 & 0.0910 & 0.1489 \\

\midrule

\textbf{\model (Ours)}
& \textbf{0.1222}
& \textbf{0.8825}
& \textbf{0.3636}
& \textbf{0.1935}
& 0.0979
& \textbf{0.1363}
& \textbf{0.1425}
& \textbf{0.8757}
& \textbf{0.2802}
& \textbf{0.0797}
& \textbf{0.0436}
& \textbf{0.1248} \\

\bottomrule
\end{tabular}
\end{table*}

%% file: sec/6_analysis.tex
\section{Analysis}
\label{sec:analysis}

We analyze the contributions of field-specific reasoning
attribution, image-quality representation, and
quality-conditioned fusion to \model to show the effectiveness of our design choices.

\subsection{Field-Specific Reasoning Attribution}
\label{sec:analysis:attribution}

\input{tables/reasoning_attribution}

We first compare field-specific reasoning with the full generation
trajectory while keeping the confidence model unchanged.
As shown in Table~\ref{tab:reasoning_attribution},
field-specific reasoning consistently improves AURC and AUROC
across both datasets and VLMs. On CUAD, it reduces AURC
by 58.9\% with Gemma and 57.5\% with Nemotron,
with similar gains on CORD. It also improves
high-precision coverage: on Gemma, Cov@98 increases
by 83.29 and 17.60 percentage points on CORD and CUAD,
respectively.
This significant improvement suggests that generation-level uncertainty
can obscure differences in reliability among fields,
whereas field-specific reasoning provides more
informative confidence signals.

\subsection{Image Quality Representation}
\label{sec:analysis:quality}

We next examine how the choice of image-quality representation
affects confidence estimation. We compare five configurations:
no quality features, a single sharpness metric following
ExtractConf~\cite{extractconf}, six global metrics,
384 tile-level measurements, and \model's combined
390-dimensional representation. All quality-aware variants
use FiLM conditioning.

\begin{figure}[t]
    \centering
    \includegraphics[width=0.90\columnwidth]
    {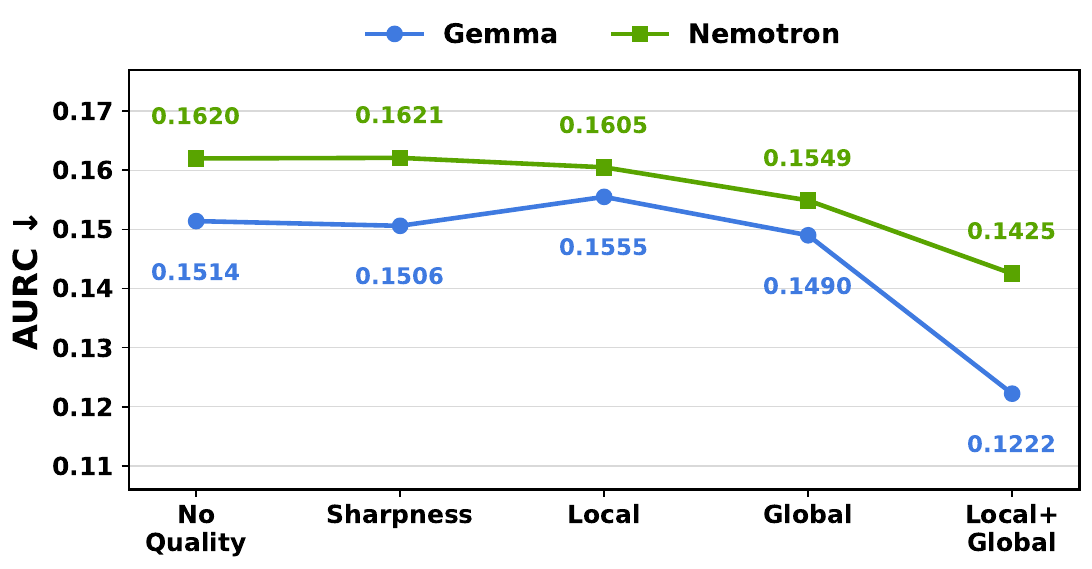}
    \caption{Effect of image-quality representations
    on CUAD. We compare no explicit quality features,
    a single sharpness metric, six global metrics,
    384 tile-level measurements, and their combined
    390-dimensional representation. Lower AURC is better.}
    \label{fig:quality_ablation}
\end{figure}

Figure~\ref{fig:quality_ablation} presents the CUAD results. The combination local and global features 
consistently achieves the strongest performance across
both VLMs. This suggests that global degradation
characteristics and spatial quality variations provide
complementary information, highlighting the benefit
of modeling image quality across multiple spatial scales.

\subsection{Image Quality Integration}
\label{sec:analysis:fusion}

\input{tables/quality_fusion}

Finally, we examine how image-quality information should
be incorporated into the confidence model. We compare
direct feature concatenation with FiLM-based conditioning,
using the same 390-dimensional local--global quality
representation. Table~\ref{tab:quality_fusion} reports
results pooled across CORD and CUAD.

Compared with direct fusion, FiLM reduces AURC by
9.4\% on Gemma and 10.9\% on
Nemotron. It also improves Cov@98 by 10.1 and
15.5 percentage points, respectively. These consistent
gains suggest that conditioning visual representations
on image quality is more effective for selective
prediction than treating quality measurements as
additional classifier inputs.





\subsection{Real-World Financial Applications}
\label{sec:analysis:production}

To assess the practical applicability of \model,
we further evaluate it on example real-world production-scale
data representative of workflows that may be typical
of a large financial firm. The evaluation covers
document processing tasks in financial services,
where accurate field extraction is important for
supporting reliable downstream operations.
Due to the regulated
nature of the data, we do not disclose the underlying
documents or detailed dataset characteristics. 
We report aggregate evaluation results to assess
the effectiveness of confidence estimation in
these real-world settings. 

\input{tables/production_results}

Table~\ref{tab:production_results} compares \model
with probability-based estimators and a learned baseline.
\model outperforms the learned baseline by 6.8\% in AUROC and 10.6\% in Cov@95,
while probability-based methods show limited discrimination. These results
highlight the potential of field-specific reasoning
and quality-aware confidence estimation in
production-scale financial document processing,
enabling greater automation while supporting
reliable downstream operations through more
effective identification of uncertain extractions.

%% file: tables/reasoning_attribution.tex
\begin{table*}[t]
\centering
\caption{Effect of field-specific reasoning attribution.
Full-trajectory reasoning uses the complete generation
trajectory, whereas field-specific reasoning uses the
reasoning attributed to each extracted field.
Best results within each dataset and VLM are in bold.}
\label{tab:reasoning_attribution}

\small
\setlength{\tabcolsep}{5pt}
\renewcommand{\arraystretch}{1.12}
\scalebox{0.80}{
    \begin{tabular}{l|cccc|cccc}
\toprule
& \multicolumn{4}{c|}{\textbf{Gemma 26B}}
& \multicolumn{4}{c}{\textbf{Nemotron 3.5 Super VL 120B}} \\
\cmidrule(lr){2-5}
\cmidrule(lr){6-9}
\textbf{Reasoning}
& AURC$\downarrow$
& AUROC$\uparrow$
& Cov@98$\uparrow$
& Brier$\downarrow$
& AURC$\downarrow$
& AUROC$\uparrow$
& Cov@98$\uparrow$
& Brier$\downarrow$ \\
\midrule

\rowcolor{gray!15}
\multicolumn{9}{c}{\textbf{CORD}} \\
\midrule

Full Trajectory
& 0.0566 & 0.5395 & 0.0355 & 0.0631
& 0.0792 & 0.5662 & 0.0181 & 0.0834 \\

\textbf{Field-Specific (Ours)}
& \textbf{0.0166}
& \textbf{0.8736}
& \textbf{0.8684}
& \textbf{0.0387}
& \textbf{0.0187}
& \textbf{0.8907}
& \textbf{0.6900}
& \textbf{0.0460} \\

\midrule

\rowcolor{gray!15}
\multicolumn{9}{c}{\textbf{CUAD}} \\
\midrule

Full Trajectory
& 0.2976 & 0.6034 & 0.0175 & 0.2303
& 0.3350 & 0.5450 & 0.0172 & 0.2332 \\

\textbf{Field-Specific (Ours)}
& \textbf{0.1222}
& \textbf{0.8825}
& \textbf{0.1935}
& \textbf{0.1363}
& \textbf{0.1425}
& \textbf{0.8757}
& \textbf{0.0797}
& \textbf{0.1248} \\

\bottomrule
\end{tabular}

}

\end{table*}

%% file: tables/quality_fusion.tex
\begin{table}[t]
\centering
\caption{Comparison of image-quality integration
strategies on pooled CORD and CUAD results.
Lower AURC and higher Cov@98 are better.}
\label{tab:quality_fusion}
\small
\setlength{\tabcolsep}{4pt}
\renewcommand{\arraystretch}{1.1}

\scalebox{0.90}{
    
\begin{tabular}{l|cc|cc}
\toprule
& \multicolumn{2}{c|}{\textbf{Gemma 26B}}
& \multicolumn{2}{c}{\textbf{Nemotron 3.5 120B}} \\
\cmidrule(lr){2-3}
\cmidrule(lr){4-5}
\textbf{Method}
& AURC$\downarrow$ & Cov@98$\uparrow$
& AURC$\downarrow$ & Cov@98$\uparrow$ \\
\midrule
Direct Fusion
& 0.0320 & 0.4900
& 0.0460 & 0.2120 \\
\textbf{FiLM (Ours)}
& \textbf{0.0290} & \textbf{0.5910}
& \textbf{0.0410} & \textbf{0.3670} \\
\bottomrule
\end{tabular}

}
\end{table}

%% file: tables/production_results.tex
\begin{table}[t]
    \centering
    \caption{Confidence estimation on private financial
    production data. AUROC and Cov@95 are reported
    as percentages; higher is better.}
    \label{tab:production_results}
    \small
    \setlength{\tabcolsep}{6pt}
    \scalebox{0.88}{
     \begin{tabular}{lcc}
        \toprule
        \textbf{Method} & \textbf{AUROC} $\uparrow$
                        & \textbf{Cov@95} $\uparrow$ \\
        \midrule
        Answer LogProb    & 54.5 & -- \\
        Reasoning LogProb & 67.1 & -- \\
        Learned Baseline  & 91.3 & 74.5 \\
        \midrule
        \textbf{\model}       & \textbf{98.1} & \textbf{85.1} \\
        \bottomrule
    \end{tabular}
    
    }
   
    \vspace{-2mm}
\end{table}

%% file: sec/7_conclusion.tex
\section{Conclusion}
\label{sec:conclusion}

We presented \model, a field-level confidence estimation
framework for reasoning-based visual document extraction.
By attributing reasoning uncertainty to individual fields
and integrating local and global image-quality measurements
through quality-conditioned visual representations, \model
captures complementary signals of extraction reliability.
We also introduced \data, a benchmark spanning receipt
and legal document extraction with two reasoning VLMs.
Experiments demonstrate consistent improvements in
confidence discrimination, selective prediction, and
probabilistic accuracy over existing estimators.
Evaluations on real-world financial production data further
demonstrate its practical potential, achieving greater
automatic coverage at high precision than an existing
production baseline. Together, these results highlight
the value of combining field-specific reasoning and
image-quality awareness to reduce human review
requirements while supporting more reliable document
automation in high-stakes applications.

%% file: sec/X_suppl.tex
\clearpage
\setcounter{page}{1}
\maketitlesupplementary
\setcounter{section}{0}
\renewcommand{\theHsection}{supp.\arabic{section}}

\section{Implementation Details}
\label{sec:method:impl}

\subsection{Architecture and Hyperparameters}
\label{sec:supp:impl_arch}

We use a reasoning VLM to generate three completions per page and record
the top $k=20$ token log probabilities for each completion. These are
used to compute the token-level uncertainty scores in
Equation~\eqref{eq:token_conf}. Each field's uncertainty sequence is
truncated or padded to 128 tokens from the tail and encoded using a GRU
with a 128-dimensional output. Reasoning text is embedded with a
pretrained BGE-M3 encoder~\cite{bge_m3}, while the visual branch uses
a frozen base SigLIP2 encoder~\cite{siglip2} and the 390-dimensional
\texttt{tile+6g} quality vector.

The sequence, text, image, and fusion mapping widths are set to 128,
the summary-statistic branch width to 64, and the quality-conditioner
hidden width to 32. We use dropout of 0.3 and train the confidence
network with AdamW~\cite{adamw}, a learning rate of $2\times10^{-3}$, batch size
of 512, and weight decay of $10^{-4}$. The checkpoint with the best
calibration coverage is selected.

The sequence encoder is a single-layer bidirectional GRU with 64 hidden
units per direction, and its final forward and backward hidden states are
concatenated into the 128-dimensional representation
$\mathbf{h}_{\mathrm{seq}}^{f}$. Reasoning excerpts are encoded into
1024-dimensional dense BGE-M3 embeddings with a maximum input length of
2048 tokens, and the page image is encoded into the 768-dimensional
pooled SigLIP2 (\texttt{siglip2-base-patch16-naflex}) embedding.
The FiLM conditioner in Equation~\eqref{eq:image_film} is a two-layer
network ($390\rightarrow32\rightarrow256$, with layer normalization, ReLU,
and dropout after the first layer) whose output is split into
$\gamma(\mathbf{q}),\beta(\mathbf{q})\in\mathbb{R}^{128}$. Its final layer
is zero-initialized, so the visual representation is unmodulated at the
start of training.

\subsection{Optimization and Data Handling}
\label{sec:supp:impl_opt}

The positive-class weight of the binary cross-entropy loss is set to the
ratio of incorrect to correct fields in the training split. Each model is
trained for at most 50 epochs with the default linear learning-rate decay
of the HuggingFace \texttt{Trainer}, bf16 mixed precision, and evaluation
after every epoch. All experiments run on a single NVIDIA A10G GPU.

Image, quality, and text features are precomputed and cached, and all
features are standardized using training-split statistics only.
Data splits are grouped by document to prevent fields from the same
document appearing in multiple splits. The calibration split is
reserved for fitting the calibration parameters and selecting the
decision threshold.

We evaluate our quality-conditioned FiLM architecture against simpler
confidence models, alternative image representations, and different
quality-injection strategies in Section~\ref{sec:experiments}.

\section{Extraction Prompt and Output Schema}
\label{sec:supp:schema}

For each page, the VLM is queried once with the prompt in
Listing~\ref{lst:supp:prompt}. The model first reasons over the document
image and then emits a single JSON object constrained by vLLM structured
decoding~\cite{vllm_structured_outputs} to the schema in
Listing~\ref{lst:supp:schema}. For each target field, the object records
the extracted value together with reasoning excerpts copied verbatim from
the preceding reasoning trace, which allows confidence features to be
attributed to individual fields within a single generation.

\subsection{Prompt Template}
\label{sec:supp:prompt}

Listing~\ref{lst:supp:prompt} shows the Jinja2 prompt template shared by
both datasets. The template is populated with the domain, document type,
field names, field descriptions, and list-field indicators of the
domain-specific extraction schema in Section~\ref{sec:supp:fields}, and
with the JSON output format in Listing~\ref{lst:supp:schema}. Beyond
specifying the output, the prompt asks the model to verify each candidate
value against the field description and to explicitly state in its
reasoning when a value is ambiguous, partially visible, or affected by low
image quality, so that such uncertainty can surface in the field-specific
reasoning excerpts.

\begin{lstlisting}[caption={VLM reasoning extraction prompt template.},label={lst:supp:prompt}]
## Task

You are a {{ domain }} document extraction system specializing in {{ doc_type }} documents.

I have provided an image of a document. Extract the requested fields by analyzing the image directly.

Your task is to extract {{ field_names }}. First reason through each field, then return your extractions along with reasoning as a single structured JSON object.

## Response Structure

Reason step by step about the document before answering. If the image quality is low or you cannot locate any value of a field, explicitly express that too in your reasoning. Then return ONLY a single JSON object matching this exact schema:

```json
{{ output_json_format }}
```

The `extractions` array MUST contain exactly one entry per requested field ({{ field_names }}).

For each entry:
- **field**: the field name, exactly as listed above.
- **reasoning**: an array of verbatim reasoning substring for the extracted value. Your reasoning trace about the field is often scattered across multiple places -- add one array entry per relevant excerpt. Each entry MUST BE AN EXACT, VERBATIM SUBSTRING of your own reasoning trace above -- copy the relevant continuous substrings character-by-character and word-for-word, including line returns. Do NOT regenerate, paraphrase, summarize, or invent new wording here.
- **value**: the extracted value as a string. For list fields, provide a JSON array string (e.g. "["a", "b"]"). If a field cannot be confidently extracted, set the value to "UNKNOWN".

## Extraction Guidelines
{%- for field in fields %}

### {{ field.name }}
{{ field.description }}
{%- if field.is_list %}

This field can have MULTIPLE values -- set "value" to a JSON array string containing every matching value (use "[]" if none are found).
{%- endif %}
{%- endfor %}

## Extraction Instructions

Extract all {{ fields|length }} fields from the document image.

For each field, follow this verification process while reasoning:
1. Locate the candidate text in the document layout and identify the label or region it appears in.
2. Check the candidate value against the field's extraction rules and description above. If it violates a rule, reject it.
3. Note any problems: ambiguity between multiple candidates, partial visibility, low image quality, or mismatch with expected format.

Now reason through the document following the steps above, then return the single JSON object described above. The `reasoning` for each field must be copied verbatim from your reasoning trace.
\end{lstlisting}

\paragraph{Aligning excerpts with token probabilities.}
To associate each excerpt with its token probabilities, we concatenate
the decoded output tokens of the completion, locate each excerpt as an
exact substring, and map its character span to the covering token
indices. The union of these indices defines the field-specific reasoning
span, over which the top-$k$ log probabilities are collected to form the
uncertainty sequence $\mathbf{c}^{f}$. Fields whose excerpts cannot be
located verbatim, for example because the model paraphrased its
reasoning, receive no token-level features and are excluded from the
confidence datasets. For Gemma 26B, this removes 16.9\% of the
field-level examples on CUAD and 6.4\% on CORD.

\begin{lstlisting}[caption={Schematic structure of the constrained
extraction output.},label={lst:supp:schema}]
[
  {
    "field": "<field name>",
    "value": "<extracted value | null>",
    "reasoning": [
      "<verbatim excerpt from the reasoning trace>",
      ...
    ]
  ...
  }
]
\end{lstlisting}

\section{Image Quality Metrics}
\label{sec:supp:quality_metrics}

\subsection{Local and Global Representation}
\label{sec:supp:quality_repr}

All quality metrics are computed on the grayscale page image after
resizing its longer side to 1024 pixels with Lanczos resampling.
For the local representation, the normalized image is divided into an
$8\times8$ grid of equal-sized tiles, and the six metrics in
Table~\ref{tab:supp:quality_metrics} are computed for each tile in
row-major order. Tiles smaller than $16\times16$ pixels are left
undefined and set to zero after standardization. The six metrics
computed on the full page are appended to obtain the 390-dimensional
vector $\mathbf{q}$ in Equation~\eqref{eq:quality_feat}. The
\emph{Sharpness Only} variant in Section~\ref{sec:analysis:quality}
uses the page-level frequency-domain sharpness alone.

\subsection{Metric Definitions}
\label{sec:supp:quality_defs}

Table~\ref{tab:supp:quality_metrics} defines the six metrics. All are
no-reference measures computed directly from pixel intensities, and they
characterize complementary aspects of degradation: edge acutance,
frequency sharpness, and perceptual blur capture defocus and motion blur,
entropy and contrast capture faded or low-contrast content, and the noise
level captures sensor and compression noise.

\begin{table}[t]
\centering
\caption{Image quality metrics computed for each tile and for the full page.}
\label{tab:supp:quality_metrics}
\small
\setlength{\tabcolsep}{3pt}
\begin{tabular}{p{0.27\columnwidth}p{0.66\columnwidth}}
\toprule
\textbf{Metric} & \textbf{Definition} \\
\midrule
Edge acutance & Mean Sobel gradient magnitude over the strongest 5\% of pixels, reflecting the steepness of edges. \\
\midrule
Frequency sharpness & Fraction of Fourier coefficients whose magnitude exceeds $1/1000$ of the maximum~\cite{fft_sharpness}. \\
\midrule
Entropy & Shannon entropy of the 256-bin intensity histogram. \\
\midrule
Perceptual blur & No-reference blur score in $[0,1]$ comparing intensity variations before and after a 9-pixel box blur~\cite{crete_blur}. \\
\midrule
Contrast & Standard deviation of pixel intensities. \\
\midrule
Noise level & Fast noise standard-deviation estimate based on a Laplacian-difference kernel~\cite{immerkaer_noise}. \\
\bottomrule
\end{tabular}
\end{table}

\section{Extraction Fields}
\label{sec:supp:fields}

Tables~\ref{tab:supp:cuad_fields} and~\ref{tab:supp:cord_fields} list
the target fields of \data together with the matching criterion used
for correctness labeling in Section~\ref{sec:supp:labeling}.

\subsection{CUAD}
\label{sec:supp:fields_cuad}
CUAD provides 41 clause categories annotated at the contract level.
We retain the eight categories whose answers are values or entities read
from the document, and discard the remaining 33 categories, whose answers
are binary clause-presence labels. As the extraction is performed page by
page, we render each contract page at 150 DPI and locate each annotated
evidence span in the page text, using exact substring matching with a
RapidFuzz partial-ratio fallback (threshold 88). The answer of a category
is then attached to every page on which its evidence appears.

\begin{table}[t]
\centering
\caption{CUAD extraction fields and matching criteria.}
\label{tab:supp:cuad_fields}
\small
\setlength{\tabcolsep}{3pt}
\begin{tabular}{p{0.30\columnwidth}p{0.47\columnwidth}l}
\toprule
\textbf{Field} & \textbf{Definition} & \textbf{Match} \\
\midrule
Document Name & Title of the contract & Text \\
\midrule
Parties & Parties who signed the contract & Parties \\
\midrule
Agreement Date & Date on which the contract was signed & Date \\
\midrule
Effective Date & Date on which the contract takes effect & Date \\
\midrule
Expiration Date & Date on which the initial term expires & Date \\
\midrule
Renewal Term & Renewal term after the initial term expires & Term \\
\midrule
Notice Period to Terminate Renewal & Notice required to terminate renewal & Term \\
\midrule
Governing Law & Jurisdiction whose law governs the contract & Gov.\ law \\
\bottomrule
\end{tabular}
\end{table}

\subsection{CORD}
\label{sec:supp:fields_cord}
We flatten the hierarchical CORD v2 annotations into 24 fields covering
purchased items, voided items, subtotals, and totals. Item-level fields
are list-valued, with one value per receipt line. We exclude the binary
tax indicator (\texttt{menu-vatyn}) and the nested sub-item fields
(\texttt{menu-sub*}), whose annotations do not correspond to
value-level extraction targets.

\begin{table}[t]
\centering
\caption{CORD extraction fields and matching criteria. Field names follow
the CORD v2 schema.}
\label{tab:supp:cord_fields}
\footnotesize
\setlength{\tabcolsep}{3pt}
\begin{tabular}{lp{0.32\columnwidth}l}
\toprule
\textbf{Field} & \textbf{Definition} & \textbf{Match} \\
\midrule
\multicolumn{3}{l}{\textit{Purchased items (list-valued)}} \\
\texttt{menu-nm} & Item name & Text \\
\texttt{menu-num} & Item code & ID \\
\texttt{menu-unitprice} & Unit price & Numeric \\
\texttt{menu-cnt} & Quantity & Numeric \\
\texttt{menu-discountprice} & Item discount & Numeric \\
\texttt{menu-price} & Item price & Numeric \\
\texttt{menu-itemsubtotal} & Item subtotal & Numeric \\
\texttt{menu-etc} & Other item information & Text \\
\midrule
\multicolumn{3}{l}{\textit{Voided items (list-valued)}} \\
\texttt{void\_menu-nm} & Voided item name & Text \\
\texttt{void\_menu-price} & Voided item price & Numeric \\
\midrule
\multicolumn{3}{l}{\textit{Subtotal}} \\
\texttt{subtotal-price} & Subtotal amount & Numeric \\
\texttt{subtotal-discount-price} & Discount amount & Numeric \\
\texttt{subtotal-service-price} & Service charge & Numeric \\
\texttt{subtotal-othersvc-price} & Other service charge & Numeric \\
\texttt{subtotal-tax-price} & Tax amount & Numeric \\
\texttt{subtotal-etc} & Other subtotal information & Text \\
\midrule
\multicolumn{3}{l}{\textit{Total}} \\
\texttt{total-price} & Total amount & Numeric \\
\texttt{total-cash-price} & Cash paid & Numeric \\
\texttt{total-change-price} & Change returned & Numeric \\
\texttt{total-creditcard-price} & Credit card payment & Numeric \\
\texttt{total-emoney-price} & Electronic payment & Numeric \\
\texttt{total-menutype-cnt} & Number of distinct items & Numeric \\
\texttt{total-menuqty-cnt} & Total item quantity & Numeric \\
\texttt{total-etc} & Other total information & Text \\
\bottomrule
\end{tabular}
\end{table}

\section{Correctness Labeling}
\label{sec:supp:labeling}

\paragraph{Example Inclusion.}
A field in a completion forms an example when the ground truth or the
prediction is present. A field with a ground-truth value but no
prediction is labeled incorrect, and a field with a prediction
but no ground-truth value is labeled incorrect, with the
CORD-specific exception described below. When both are present, the
field-type-specific matcher determines the label.

\subsection{Value Normalization}
\label{sec:supp:label_norm}
Before matching, values are lowercased, stripped of surrounding quotes,
and whitespace-collapsed. For token-based comparisons, spelled-out numbers
(\eg, ``thirty'') are mapped to digits.

\paragraph{Matching Rules.}
\begin{itemize}
    \item \textbf{Text}: correct if the normalized strings are equal, all
    ground-truth tokens appear in the prediction, or the RapidFuzz
    token-set ratio is at least 90.
    \item \textbf{Governing law}: as Text, but token containment is
    accepted in either direction, since annotations may add or omit
    country qualifiers.
    \item \textbf{Date}: both values are parsed with month-first fuzzy
    date parsing and compared as calendar dates, accommodating formats
    such as ``8th day of May 2014'' and ``5/8/14''.
    \item \textbf{Term}: correct if the (number, unit) pairs, with units
    in years, months, weeks, or days, intersect; otherwise a token-set
    ratio of at least 90 is required.
    \item \textbf{Numeric}: currency symbols, letters, and spaces are
    removed; a comma or period followed by exactly three digits is treated
    as a thousands separator, and any remaining comma as a decimal point.
    Values are correct if numerically equal, with signs preserved, so that
    ``46.636'' equals ``46,636'' and ``2.00'' equals ``2''.
    \item \textbf{ID}: exact match after removing non-alphanumeric
    characters, without fuzzy tolerance.
    \item \textbf{Parties}: values are split on semicolons, and the field
    is correct if at least one predicted party matches a ground-truth
    party exactly or with a token-set ratio of at least 85.
\end{itemize}

\section{Additional Main-Comparison Details}
\label{sec:supp:main_comparison}

We provide the complete confidence-baseline comparisons omitted from the
main paper, together with the feature configurations and evaluation details.
All results use the same document-level partitions and field-level correctness
labels as the main experiments. Each learned confidence estimator is trained
independently for each dataset and extraction backbone; the extraction VLMs
are not retrained for these comparisons.

\subsection{Complete Baseline Results}
\label{sec:supp:baseline_results}

Tables~\ref{tab:supp:cord_full} and~\ref{tab:supp:cuad_full} extend
Table~\ref{tab:main_results} with answer-token maximum/minimum log
probability, answer entropy, answer probability margin, and a
reasoning-text-only estimator. We report all six metrics for every
available method. Values are displayed to four decimal places.

\begin{table*}[t]
\centering
\caption{Complete field-level confidence baseline comparison on CORD.
AURC, ECE, and Brier are lower-is-better; AUROC, Cov@95, and Cov@98
are higher-is-better. Best values are highlighted in bold.}
\label{tab:supp:cord_full}
\scriptsize
\setlength{\tabcolsep}{2.3pt}
\renewcommand{\arraystretch}{1.08}
\begin{tabular}{l|cccccc|cccccc}
\toprule
& \multicolumn{6}{c|}{\textbf{Gemma 26B}}
& \multicolumn{6}{c}{\textbf{Nemotron 120B}} \\
\cmidrule(lr){2-7}\cmidrule(lr){8-13}
\textbf{Method}
& AURC$\downarrow$ & AUROC$\uparrow$ & Cov@95$\uparrow$ & Cov@98$\uparrow$ & ECE$\downarrow$ & Brier$\downarrow$
& AURC$\downarrow$ & AUROC$\uparrow$ & Cov@95$\uparrow$ & Cov@98$\uparrow$ & ECE$\downarrow$ & Brier$\downarrow$ \\
\midrule

Sample Agreement & 0.0367 & 0.6648 & 0.8635 & 0.2115 & 0.0264 & 0.0641 & 0.0503 & 0.7375 & 0.7115 & 0.0156 & 0.0254 & 0.0873 \\

Answer LogProb (Mean) & 0.0544 & 0.4960 & 0.9869 & 0.0018 & \textbf{0.0179} & 0.0489 & 0.1083 & 0.4324 & 0.0000 & 0.0000 & 0.0384 & 0.0879 \\

Answer LogProb (Max) & 0.0551 & 0.5236 & \textbf{0.9967} & 0.0050 & \textbf{0.0179} & 0.0491 & 0.0808 & 0.4718 & 0.0197 & 0.0173 & 0.0376 & 0.0878 \\

Answer LogProb (Min) & 0.0550 & 0.4879 & 0.9743 & 0.0018 & \textbf{0.0179} & 0.0489 & 0.1137 & 0.4037 & 0.0000 & 0.0000 & 0.0401 & 0.0881 \\

Answer Entropy & 0.0514 & 0.4950 & 0.9888 & 0.0000 & \textbf{0.0179} & 0.0489 & 0.1160 & 0.4320 & 0.0018 & 0.0018 & 0.0390 & 0.0880 \\

Answer Margin & 0.0484 & 0.5078 & 0.9907 & 0.0029 & 0.0221 & 0.0488 & 0.1063 & 0.4389 & 0.0000 & 0.0000 & 0.0367 & 0.0858 \\

Reasoning LogProb & 0.0545 & 0.5593 & 0.2661 & 0.0419 & 0.0271 & 0.0651 & 0.1051 & 0.4923 & 0.0030 & 0.0030 & 0.0374 & 0.1008 \\

\midrule

Reasoning Trajectory & 0.0334 & 0.7316 & 0.9164 & 0.1832 & 0.0210 & 0.0583 & 0.0614 & 0.6752 & 0.4172 & 0.0355 & 0.0310 & 0.0919 \\

Reasoning Text & 0.0712 & 0.4672 & 0.0007 & 0.0007 & -- & -- & 0.1061 & 0.4386 & 0.0785 & 0.0035 & -- & -- \\

Reasoning + Text & 0.0168 & 0.8497 & 0.9774 & 0.7086 & 0.0223 & 0.0434 & 0.0342 & 0.8487 & 0.9178 & 0.0041 & 0.0307 & 0.0505 \\

\midrule

Reasoning + Visual & 0.0173 & 0.8414 & 0.9774 & 0.6468 & 0.0260 & 0.0456 & 0.0273 & 0.8592 & 0.9122 & 0.2671 & 0.0287 & 0.0558 \\

\midrule

\textbf{\model (Ours)} & \textbf{0.0166} & \textbf{0.8736} & 0.9839 & \textbf{0.8684} & 0.0188 & \textbf{0.0387} & \textbf{0.0187} & \textbf{0.8907} & \textbf{0.9539} & \textbf{0.6900} & \textbf{0.0197} & \textbf{0.0460} \\

\bottomrule
\end{tabular}
\end{table*}

\begin{table*}[t]
\centering
\caption{Complete field-level confidence baseline comparison on CUAD.
AURC, ECE, and Brier are lower-is-better; AUROC, Cov@95, and Cov@98
are higher-is-better. Best values are highlighted in bold.}
\label{tab:supp:cuad_full}
\scriptsize
\setlength{\tabcolsep}{2.3pt}
\renewcommand{\arraystretch}{1.08}
\begin{tabular}{l|cccccc|cccccc}
\toprule
& \multicolumn{6}{c|}{\textbf{Gemma 26B}}
& \multicolumn{6}{c}{\textbf{Nemotron 120B}} \\
\cmidrule(lr){2-7}\cmidrule(lr){8-13}
\textbf{Method}
& AURC$\downarrow$ & AUROC$\uparrow$ & Cov@95$\uparrow$ & Cov@98$\uparrow$ & ECE$\downarrow$ & Brier$\downarrow$
& AURC$\downarrow$ & AUROC$\uparrow$ & Cov@95$\uparrow$ & Cov@98$\uparrow$ & ECE$\downarrow$ & Brier$\downarrow$ \\
\midrule

Sample Agreement & 0.3998 & 0.4640 & 0.0076 & 0.0076 & \textbf{0.0902} & 0.2499 & 0.3842 & 0.5067 & 0.0038 & 0.0038 & 0.0898 & 0.2467 \\

Answer LogProb (Mean) & 0.2464 & 0.4292 & 0.0158 & 0.0090 & 0.1175 & 0.1824 & 0.3506 & 0.3877 & 0.0042 & 0.0042 & 0.0564 & 0.2024 \\

Answer LogProb (Max) & 0.2450 & 0.4921 & 0.0554 & 0.0501 & 0.1175 & 0.1824 & 0.2762 & 0.5032 & 0.0127 & 0.0127 & 0.0564 & 0.2025 \\

Answer LogProb (Min) & 0.2666 & 0.4024 & 0.0158 & 0.0090 & 0.1175 & 0.1824 & 0.3641 & 0.3718 & 0.0042 & 0.0042 & 0.0564 & 0.2027 \\

Answer Entropy & 0.2511 & 0.4253 & 0.0000 & 0.0000 & 0.1175 & 0.1824 & 0.3492 & 0.3888 & 0.0021 & 0.0021 & 0.0564 & 0.2024 \\

Answer Margin & 0.2335 & 0.3874 & 0.0170 & 0.0170 & 0.1175 & 0.1844 & 0.3581 & 0.3870 & 0.0000 & 0.0000 & 0.0564 & 0.2027 \\

Reasoning LogProb & 0.3057 & 0.6431 & 0.0000 & 0.0000 & 0.1494 & 0.2323 & 0.3743 & 0.4857 & 0.0000 & 0.0000 & 0.1164 & 0.2552 \\

\midrule

Reasoning Trajectory & 0.2162 & 0.7316 & 0.0095 & 0.0095 & 0.1689 & 0.2201 & 0.2782 & 0.6150 & 0.0129 & 0.0128 & 0.1142 & 0.2390 \\

Reasoning Text & 0.3117 & 0.3778 & 0.0050 & 0.0050 & -- & -- & 0.3283 & 0.3767 & 0.0000 & 0.0000 & -- & -- \\

Reasoning + Text & 0.1557 & 0.8662 & 0.2010 & 0.0327 & 0.1225 & 0.1434 & 0.1468 & 0.8649 & 0.0903 & 0.0186 & 0.0739 & 0.1281 \\

\midrule

Reasoning + Visual & 0.1514 & 0.8684 & 0.2397 & 0.0227 & 0.1341 & 0.1621 & 0.1620 & 0.8429 & 0.0456 & 0.0143 & 0.0910 & 0.1489 \\

\midrule

\textbf{\model (Ours)} & \textbf{0.1222} & \textbf{0.8825} & \textbf{0.3636} & \textbf{0.1935} & 0.0979 & \textbf{0.1363} & \textbf{0.1425} & \textbf{0.8757} & \textbf{0.2802} & \textbf{0.0797} & \textbf{0.0436} & \textbf{0.1248} \\

\bottomrule
\end{tabular}
\end{table*}

\subsection{Baseline Definitions and Feature Configurations}
\label{sec:supp:baseline_defs}

\paragraph{Training-Free Confidence Scores.}
\textit{Sample Agreement} uses three independently generated completions
of the same document and computes agreement for each target field; unlike
single-completion methods, it requires three extraction generations.
\textit{Answer LogProb (Mean/Max/Min)} aggregates the token log
probabilities over the tokens corresponding to the extracted field answer.
\textit{Answer Entropy} aggregates token-level predictive uncertainty,
while \textit{Answer Margin} uses the separation between the leading
token candidates. \textit{Reasoning LogProb} aggregates the field-aligned
reasoning-token probabilities, without training a classifier.

Specifically, Sample Agreement is the fraction of the three completions
whose normalized value for the field equals that of the current
completion. The answer tokens are located as the last occurrence of the
extracted value in the decoded output, since the answer JSON follows the
reasoning trace. Answer LogProb takes the mean, maximum, or minimum of the
sampled-token log probabilities over these tokens. Answer Entropy
averages, over the answer tokens, the Shannon entropy of the top-$k$
distribution renormalized to sum to one, and Answer Margin averages the
log-probability gap between the two most likely candidates. Reasoning
LogProb is the mean sampled-token log probability over the field-specific
reasoning span. All scores are oriented so that higher values indicate
higher confidence, with entropy negated.

\paragraph{Learned Confidence Estimators.}
\textit{Reasoning Trajectory} uses a GRU representation of the final $L$
reasoning-token uncertainty values together with twelve summary statistics.
\textit{Reasoning Text} uses the pretrained BGE-M3 embedding of the
field-specific reasoning excerpt. \textit{Reasoning + Text} combines
these two branches. \textit{Reasoning + Visual} adds the pretrained SigLIP2
page embedding but excludes explicit image-quality conditioning.
\model adds the 390-dimensional image-quality descriptor, comprising
six metrics for each of 64 spatial tiles and six page-level metrics,
and uses FiLM to modulate the visual embedding before feature fusion.
All learned methods use an independent confidence predictor with only
the corresponding feature branches enabled.

\begin{table}[t]
\centering
\caption{Input configurations of learned confidence estimators.
``Seq.'' denotes GRU-based uncertainty sequence features, ``Stats''
denotes twelve summary statistics, and ``Text'' denotes BGE-M3 embeddings.}
\label{tab:supp:baseline_features}
\small
\setlength{\tabcolsep}{3.5pt}
\begin{tabular}{lccccc}
\toprule
\textbf{Method} & Seq. & Stats & Text & SigLIP2 & FiLM \\
\midrule
Reasoning Trajectory  & $\checkmark$ & $\checkmark$ & -- & -- & -- \\
Reasoning Text        & -- & -- & $\checkmark$ & -- & -- \\
Reasoning + Text      & $\checkmark$ & $\checkmark$ & $\checkmark$ & -- & -- \\
Reasoning + Visual    & $\checkmark$ & $\checkmark$ & $\checkmark$ & $\checkmark$ & -- \\
\midrule
\textbf{\model}    & $\checkmark$ & $\checkmark$ & $\checkmark$ & $\checkmark$ & $\checkmark$ \\
\bottomrule
\end{tabular}
\end{table}

\subsection{Training, Calibration, and Evaluation Protocol}
\label{sec:supp:main_protocol}

\paragraph{Partitioning and Training.}
All configurations share the same document-level training,
calibration, and test partition within each dataset--VLM setting.
Predictions from independent completions of a source document are kept
in the same partition to prevent leakage. Learned baselines are fitted
independently with matched optimization and early-stopping procedures;
pretrained text and visual encoders remain frozen as specified in the
main method. Training-free confidence scores do not require a learned
classification network.
Documents are randomly assigned to the training, calibration, and test
partitions in an 80/10/10 ratio, yielding the example counts in
Table~\ref{tab:data_stats}.

\paragraph{Calibration.}
For learned estimators, scalar temperature and bias parameters are fitted
on the held-out calibration partition using negative log likelihood:
\begin{equation}
\hat p^f = \sigma(s^f/T + b_{\mathrm{cal}}), \qquad T>0.
\end{equation}
The training-free scores are mapped to probabilities with the same affine
transformation, fitted on the same calibration partition, before computing
ECE and Brier. As this mapping is monotonic, it leaves AURC, AUROC, and
coverage unchanged.